# The Pros and Cons of Using Machine Learning and Interpretable Machine Learning Methods In Psychiatry Detection Applications, Specifically Depression Disorder: A Brief Review.

Hossein Simchi[1], Samira Tajik[1]

1- Faculty of Computer Science and Engineering, Shahid Beheshti University, Tehran, Iran.

Corresponding Email: h.simchi@alumni.sbu.ac.ir

## Abstract

The COVID-19 pandemic has forced many people to limit their social activities, which has resulted in a rise in mental illnesses, particularly depression. To diagnose these illnesses with accuracy and speed, and prevent severe outcomes such as suicide, the use of machine learning has become increasingly important. Additionally, to provide precise and understandable diagnoses for better treatment, AI scientists and researchers must develop interpretable AI-based solutions. This article provides an overview of relevant articles in the field of machine learning and interpretable AI, which helps to understand the advantages and disadvantages of using AI in psychiatry disorder detection applications.

# 1- Introduction

Maintaining good mental health is essential for an individual's overall well-being. It has a significant impact on the quality of life and should be given utmost importance. Therefore, it is crucial to understand the nature of mental illness and treat it as early as possible.

Moreover, medical practitioners recognize the importance of mental health in an individual's overall well-being and its impact on society. They strive to identify and treat illnesses early with high accuracy. Studies have shown that the COVID-19 pandemic has led to a significant increase in mental health issues, particularly anxiety and depression[1-4]. The emergence of this disease has significantly increased the prevalence of anxiety, depression, post-traumatic stress disorder (PTSD), psychological distress, and stress in some countries around the world [5].

In addition, machine learning has played an increasingly important role in medical applications due to its high speed and accuracy [6, 7]. However, the use of machine learning in mental illnesses has not been fully investigated. It remains to be answered whether machine learning can be useful in detecting mental illnesses for clinical purposes.

Also, ethical considerations must be taken into account when using machine learning. These include bias, transparency, and explainability[8, 9]. First, to train machine learning models, it is crucial to carefully select data to ensure that the model doesn't overfit. Second, patients and physicians require clear and transparent explanations for the model's output, leading to interpretable artificial intelligence (XAI).

Finally, this study aims to review the pros and cons of using machine learning and interpretable machine learning methods in detecting mental disorders, specifically depression.

The article includes the following section:

- Section 2 explains the use of machine learning for diagnosing mental illnesses.

Notably, to write a useful review article, the study has been selected recent articles and relevant topics.

## 2- The use of machine learning in diagnosing mental illnesses

Machine learning algorithms can be classified into two categories: traditional and deep learning algorithms. In traditional algorithms, feature engineering is manually performed using different methods such as random forest. In contrast, deep learning-based methods automate feature engineering using deep neural networks, which increases both speed and accuracy. Due to the high speed and accuracy of machine learning algorithms, particularly those based on deep learning, many neural network-based approaches have been used to diagnose psychiatric disorders in recent years [10-18]. There are various applications of machine learning in the diagnosis and treatment of psychiatric disorders The following are some benefits of using machine learning in the diagnosis and treatment of psychiatric disorders [19-22] :

1. Creation of smart diagnosis tools with high accuracy.

2. Monitoring of symptoms: physiological data such as heart rate variability and sleep quality can be collected to identify risk factors and provide personalized care.

3. Delivery of personalized treatment recommendations.

4. Detailed explanation of the reasoning behind the desired treatment for the patient.

5. Visualization of the process steps, which can be analyzed to make more accurate decisions in proceeding with treatment steps.

6. Early diagnosis, which can reduce the side effects of the disease and prevent suicide in some types of mental illness such as depression.

7. Improved long-term disease management, since using artificial intelligence to check the treatment steps will minimize errors in the diagnosis and treatment process over time.

8. The ability to explain why the disease occurs and why a particular drug is used to treat it.

## 2-1- Machine learning methods

Over the years, numerous neural network-based approaches have been developed for diagnosing mental disorders such as anxiety disorder, schizophrenia [23-28] , bipolar disorder [23, 25, 29], post-traumatic stress disorder (PTSD) [23], anorexia nervosa disorder [23], attention-deficit hyperactivity disorder (ADHD) [23, 30, 31], mood analysis [32, 33], suicide risk management[34-37], obsessive-compulsive disorder (OCD) [38], and autism spectrum disorder (ASD)[39]. Meanwhile, Electroencephalogram (EEG) data, which records the electrical activity of the brain, has been widely used to test many of these methods and investigate various neurological conditions including epilepsy, sleep disorders, and brain injuries[40]. Eventually, There are many studies to suggest that EEG data holds potential for detecting mental illnesses such as depression disorder[41-49].

In recent years, there have been a few studies on predicting anxiety disorders. One such study is the X-AI model introduced by [50] , which predicts anxiety in elderly individuals who live alone in South Korea. The study analyzed 1,558 elderly individuals to determine the factors that predict anxiety disorders. Another study by Priya et al [51] compared the performance of five machine learning algorithms in detecting anxiety, stress, and depression. After evaluating different measures such as accuracy, error rate, precision, recall, specificity, and F1-score, they found that the Random Forest classifier had the highest F1-score for detecting anxiety and stress. They also reported that Naïve Bayes achieved the highest F1-score and accuracy for depression. DeprNet developed a deep convolutional neural network framework for detecting depression using EEG data [52]. They also proposed two brain-computer interfaces (BCIs) for clinical use to detect the levels of depression and anxiety.

Also, depression is a serious mental disorder[53], and various methods have been developed to detect it. These include text processing, image processing, emotional chatbots, and sentiment analysis [54-57]. However, it is essential to identify its symptoms before attempting to detect it. Therefore, several studies have been conducted to identify these symptoms [58, 59]. When it comes to detecting depression, it is crucial to identify warning signs for suicide. To address this issue, Zohuri et al. [60] conducted a study to explore the potential and limitations of machine learning in analyzing mood, detecting depression, and managing suicide risk. In the study of [61], machine learning models, specifically KNN and AdaBoost, were used to predict depression in patients with Parkinson's disease. The researchers used an ensemble model to combine the predictions of multiple machine learning models, which improved the accuracy of the prediction. In another study, Vanhollebeke et al. [62], introduced a practical tool for diagnosing depression that uses three different feature types extracted from brain regions. They trained an SVM model on these features to understand which features are important in detecting depression.

Moreover, With the development of technologies such as smartphones and social media applications, numerous studies have highlighted their potential to track and detect mental illnesses, particularly depression[53, 63-74]. For instance, in a study by [70], researchers used GPS traces of people's mobility patterns to monitor individuals affected by depression mood disorder. Researchers have also been analyzing the use of facial images and videos for detecting depression[75]. One such study involves a framework called DepressNet, which aims to automate depression diagnosis (ADD) using facial images. The researchers used a deep convolutional neural network (CNN) to train on a visual depression dataset, which helped identify the significant areas of input images in terms of their severity score. This was done by generating a depression activation map (DAM)[76].

## 2-2- Interpretable machine learning methods

Machine learning-based intelligent systems are widely used to train data and predict its output. However, these systems cannot provide a reason for their predictions [77]. Therefore, given the importance of disease type and treatment prediction in medical applications, machine learning systems will not be reliable if there is not enough reason to predict. The importance of using interpretable artificial intelligence (XAI) in medical applications can be described as follows [78]:

1- It increases the trust and acceptance of doctors and patients
2- It improves the accuracy and reliability of diagnosis
3- It helps identify and correct any errors or biases in the system
4- It ensures that the AI system makes decisions based on relevant and meaningful factors

Many studies have shown the importance of XAI in mental health and psychiatry [79-81]. For instance, Joyce et al. [82] introduced a novel framework called TIFU, which provides recommendations for evaluating AI models that are transparent, fair, and understandable.

Another study presented an interactive visualization technique called RullMatrix, which can help users with limited knowledge of machine learning understand and validate the predictions of the model [83].

## 3- Discussion and Conclusion

In aggrement with previous studies, our findings showed that the use of machine learning methods and XAI methods is necessary in medical applications specially mental illnesses. Furthermore, we have observed that the use of artificial intelligence, despite its inherent benefits, may also present ethical considerations that need to be carefully considered.

## 4- References

1. Hawes, M.T., et al., *Increases in depression and anxiety symptoms in adolescents and young adults during the COVID-19 pandemic.* Psychological medicine, 2022. **52**(14): p. 3222-3230.
2. Pfefferbaum, B. and C.S. North, *Mental health and the Covid-19 pandemic.* New England journal of medicine, 2020. **383**(6): p. 510-512.
3. Galea, S., R.M. Merchant, and N. Lurie, *The mental health consequences of COVID-19 and physical distancing: the need for prevention and early intervention.* JAMA internal medicine, 2020. **180**(6): p. 817-818.
4. Wang, X., et al., *Investigating mental health of US college students during the COVID-19 pandemic: Cross-sectional survey study.* Journal of medical Internet research, 2020. **22**(9): p. e22817.
5. Xiong, J., et al., *Impact of COVID-19 pandemic on mental health in the general population: A systematic review.* Journal of affective disorders, 2020. **277**: p. 55-64.
6. Nabavi, S., et al., *Automatic multi-class cardiovascular magnetic resonance image quality assessment using unsupervised domain adaptation in spatial and frequency domains.* arXiv preprint arXiv:2112.06806, 2021.
7. Nabavi, S., et al., *A Generalised Deep Meta-Learning Model for Automated Quality Control of Cardiovascular Magnetic Resonance Images.* arXiv preprint arXiv:2303.13324, 2023.
8. Fusar-Poli, P., et al., *Ethical considerations for precision psychiatry: A roadmap for research and clinical practice.* European Neuropsychopharmacology, 2022. **63**: p. 17-34.
9. McCradden, M., K. Hui, and D.Z. Buchman, *Evidence, ethics and the promise of artificial intelligence in psychiatry.* Journal of Medical Ethics, 2023. **49**(8): p. 573-579.
10. Li, F., et al., *Artificial intelligence applications in psychoradiology.* Psychoradiology, 2021. **1**(2): p. 94-107.

11. Davatzikos, C. and T.D. Satterthwaite, *Commentary to "Translational machine learning for child and adolescent psychiatry".* Journal of Child Psychology and Psychiatry, 2022. **63**(4): p. 444-446.
12. Graham, S., et al., *Artificial intelligence for mental health and mental illnesses: an overview.* Current psychiatry reports, 2019. **21**: p. 1-18.
13. Liu, C., et al., *Review on the Application of Artificial Intelligence in Bioinformatics.* Highlights in Science, Engineering and Technology, 2023. **30**: p. 209-214.
14. Smucny, J., G. Shi, and I. Davidson, *Deep learning in neuroimaging: Overcoming challenges with emerging approaches.* Frontiers in Psychiatry, 2022. **13**: p. 912600.
15. Zhou, S., J. Zhao, and L. Zhang, *Application of artificial intelligence on psychological interventions and diagnosis: an overview.* Frontiers in Psychiatry, 2022. **13**: p. 811665.
16. Chen, Z.S., et al., *Modern views of machine learning for precision psychiatry.* Patterns, 2022. **3**(11).
17. Chekroud, A.M., et al., *The promise of machine learning in predicting treatment outcomes in psychiatry.* World Psychiatry, 2021. **20**(2): p. 154-170.
18. Stefano, G.B., et al., *Artificial Intelligence: Deciphering the Links between Psychiatric Disorders and Neurodegenerative Disease*. 2023, MDPI. p. 1055.
19. Rocheteau, E., *On the role of artificial intelligence in psychiatry.* The British Journal of Psychiatry, 2023. **222**(2): p. 54-57.
20. Lin, B., G. Cecchi, and D. Bouneffouf, *Psychotherapy AI companion with reinforcement learning recommendations and interpretable policy dynamics.* arXiv preprint arXiv:2303.09601, 2023.
21. Delanerolle, G., et al., *Artificial intelligence: a rapid case for advancement in the personalization of gynaecology/obstetric and mental health care.* Women's Health, 2021. **17**: p. 17455065211018111.
22. Farah, L., et al., *Assessment of Performance, Interpretability, and Explainability in Artificial Intelligence–Based Health Technologies: What Healthcare Stakeholders Need to Know.* Mayo Clinic Proceedings: Digital Health, 2023. **1**(2): p. 120-138.
23. Iyortsuun, N.K., et al. *A Review of Machine Learning and Deep Learning Approaches on Mental Health Diagnosis*. in *Healthcare*. 2023. MDPI.
24. Organisciak, D., et al., *RobIn: A robust interpretable deep network for schizophrenia diagnosis.* Expert Systems with Applications, 2022. **201**: p. 117158.
25. Li, D., W. Liu, and H. Han, *Mislabeled learning for psychiatric disorder detection.* medRxiv, 2022: p. 2022.08. 11.22278675.

26. Mellem, M.S., et al., *Explainable AI enables clinical trial patient selection to retrospectively improve treatment effects in schizophrenia.* BMC medical informatics and decision making, 2021. **21**(1): p. 162.

27. Cortes-Briones, J.A., et al., *Going deep into schizophrenia with artificial intelligence.* Schizophrenia Research, 2022. **245**: p. 122-140.

28. Nsugbe, E., *Enhanced recognition of adolescents with schizophrenia and a computational contrast of their neuroanatomy with healthy patients using brainwave signals.* Applied AI Letters, 2023. **4**(1): p. e79.

29. Grünerbl, A., et al., *Smartphone-based recognition of states and state changes in bipolar disorder patients.* IEEE journal of biomedical and health informatics, 2014. **19**(1): p. 140-148.

30. Khare, S.K. and U.R. Acharya, *An explainable and interpretable model for attention deficit hyperactivity disorder in children using EEG signals.* Computers in biology and medicine, 2023. **155**: p. 106676.

31. Chen, T., et al., *Diagnosing attention-deficit hyperactivity disorder (ADHD) using artificial intelligence: a clinical study in the UK.* Frontiers in Psychiatry, 2023. **14**: p. 1164433.

32. Thosar, D., et al., *Review on mood detection using image processing and chatbot using artificial intelligence.* life, 2018. **5**(03).

33. Banerjee, T., et al., *Predicting mood disorder symptoms with remotely collected videos using an interpretable multimodal dynamic attention fusion network.* arXiv preprint arXiv:2109.03029, 2021.

34. Cansel, N., et al., *Interpretable estimation of suicide risk and severity from complete blood count parameters with explainable artificial intelligence methods.* Psychiatria Danubina, 2023. **35**(1): p. 62-72.

35. Hong, S., et al., *Identification of suicidality in adolescent major depressive disorder patients using sMRI: A machine learning approach.* Journal of affective disorders, 2021. **280**: p. 72-76.

36. Martínez-Miranda, J., *Embodied conversational agents for the detection and prevention of suicidal behaviour: current applications and open challenges.* Journal of medical systems, 2017. **41**(9): p. 135.

37. Fonseka, T.M., V. Bhat, and S.H. Kennedy, *The utility of artificial intelligence in suicide risk prediction and the management of suicidal behaviors.* Australian & New Zealand Journal of Psychiatry, 2019. **53**(10): p. 954-964.

38. Kalmady, S.V., et al., *Prediction of obsessive-compulsive disorder: importance of neurobiology-aided feature design and cross-diagnosis transfer learning.* Biological Psychiatry: Cognitive Neuroscience and Neuroimaging, 2022. **7**(7): p. 735-746.

39. Ke, F., et al., *Exploring the structural and strategic bases of autism spectrum disorders with deep learning.* Ieee Access, 2020. **8**: p. 153341-153352.

40. Salama, E.S., et al., *EEG-based emotion recognition using 3D convolutional neural networks.* International Journal of Advanced Computer Science and Applications, 2018. **9**(8).

41. Sovatzidi, G., M. Vasilakakis, and D.K. Iakovidis, *Constructive Fuzzy Cognitive Map for Depression Severity Estimation.* Studies in health technology and informatics, 2022. **294**: p. 485-489.

42. Li, X., et al., *Depression recognition using machine learning methods with different feature generation strategies.* Artificial intelligence in medicine, 2019. **99**: p. 101696.

43. Yasin, S., et al., *EEG based Major Depressive disorder and Bipolar disorder detection using Neural Networks: A review.* Computer Methods and Programs in Biomedicine, 2021. **202**: p. 106007.

44. Li, X., et al., *EEG-based mild depression recognition using convolutional neural network.* Medical & biological engineering & computing, 2019. **57**: p. 1341-1352.

45. Ay, B., et al., *Automated depression detection using deep representation and sequence learning with EEG signals.* Journal of medical systems, 2019. **43**: p. 1-12.

46. Safayari, A. and H. Bolhasani, *Depression diagnosis by deep learning using EEG signals: A systematic review.* Medicine in Novel Technology and Devices, 2021. **12**: p. 100102.

47. Arns, M., et al., *EEG alpha asymmetry as a gender-specific predictor of outcome to acute treatment with different antidepressant medications in the randomized iSPOT-D study.* Clinical Neurophysiology, 2016. **127**(1): p. 509-519.

48. Olbrich, S. and M. Arns, *EEG biomarkers in major depressive disorder: discriminative power and prediction of treatment response.* International Review of Psychiatry, 2013. **25**(5): p. 604-618.

49. Wang, Z., et al., *Automated rest eeg-based diagnosis of depression and schizophrenia using a deep convolutional neural network.* IEEE Access, 2022. **10**: p. 104472-104485.

50. Byeon, H., *Exploring factors for predicting anxiety disorders of the elderly living alone in South Korea using interpretable machine learning: A population-based study.* International Journal of Environmental Research and Public Health, 2021. **18**(14): p. 7625.

51. Priya, A., S. Garg, and N.P. Tigga, *Predicting anxiety, depression and stress in modern life using machine learning algorithms.* Procedia Computer Science, 2020. **167**: p. 1258-1267.

52. Seal, A., et al., *DeprNet: A deep convolution neural network framework for detecting depression using EEG.* IEEE Transactions on Instrumentation and Measurement, 2021. **70**: p. 1-13.

53. Giuntini, F.T., et al., *A review on recognizing depression in social networks: challenges and opportunities.* Journal of Ambient Intelligence and Humanized Computing, 2020. **11**: p. 4713-4729.
54. Joshi, M.L. and N. Kanoongo, *Depression detection using emotional artificial intelligence and machine learning: A closer review.* Materials Today: Proceedings, 2022. **58**: p. 217-226.
55. Benrimoh, D., et al., *Towards Outcome-Driven Patient Subgroups: A Machine Learning Analysis Across Six Depression Treatment Studies.* The American Journal of Geriatric Psychiatry, 2023.
56. Gao, S., V.D. Calhoun, and J. Sui, *Machine learning in major depression: From classification to treatment outcome prediction.* CNS neuroscience & therapeutics, 2018. **24**(11): p. 1037-1052.
57. Cummins, N., et al., *A review of depression and suicide risk assessment using speech analysis.* Speech communication, 2015. **71**: p. 10-49.
58. Uddin, M.Z., et al., *Deep learning for prediction of depressive symptoms in a large textual dataset.* Neural Computing and Applications, 2022. **34**(1): p. 721-744.
59. Steijn, F.v., *Speech-based Depression Prediction with Symptoms as Interpretable Intermediate Features.* 2022.
60. Zohuri, B. and S. Zadeh, *The utility of artificial intelligence for mood analysis, depression detection, and suicide risk management.* Journal of Health Science, 2020. **8**(2): p. 67-73.
61. Nguyen, H.V. and H. Byeon, *Prediction of Parkinson's Disease Depression Using LIME-Based Stacking Ensemble Model.* Mathematics, 2023. **11**(3): p. 708.
62. Vanhollebeke, G., et al. *Diagnosis of depression based on resting state functional MRI.* in *18th National Day on Biomedical Engineering: Artificial Intelligence in Medicine.* 2019. NCBME.
63. Farhan, A.A., et al. *Behavior vs. introspection: refining prediction of clinical depression via smartphone sensing data.* in *2016 IEEE wireless health (WH).* 2016. IEEE.
64. Saeb, S., et al., *Mobile phone sensor correlates of depressive symptom severity in daily-life behavior: an exploratory study.* Journal of medical Internet research, 2015. **17**(7): p. e4273.
65. Wahle, F., et al., *Mobile sensing and support for people with depression: a pilot trial in the wild.* JMIR mHealth and uHealth, 2016. **4**(3): p. e5960.
66. Wang, R., et al., *Tracking depression dynamics in college students using mobile phone and wearable sensing.* Proceedings of the ACM on Interactive, Mobile, Wearable and Ubiquitous Technologies, 2018. **2**(1): p. 1-26.
67. Xie, J., et al., *Care for the Mind Amid Chronic Diseases: An Interpretable AI Approach Using IoT.* arXiv preprint arXiv:2211.04509, 2022.

68. Song, H., et al. *Feature attention network: interpretable depression detection from social media*. in *Proceedings of the 32nd Pacific Asia conference on language, information and computation*. 2018.
69. Zainab, R. and R. Chandramouli. *Detecting and explaining depression in social media text with machine learning*. in *GOOD Workshop KDD*. 2020.
70. Canzian, L. and M. Musolesi. *Trajectories of depression: unobtrusive monitoring of depressive states by means of smartphone mobility traces analysis*. in *Proceedings of the 2015 ACM international joint conference on pervasive and ubiquitous computing*. 2015.
71. Thati, R.P., et al., *A novel multi-modal depression detection approach based on mobile crowd sensing and task-based mechanisms.* Multimedia Tools and Applications, 2023. **82**(4): p. 4787-4820.
72. De Choudhury, M., S. Counts, and E. Horvitz. *Predicting postpartum changes in emotion and behavior via social media*. in *Proceedings of the SIGCHI conference on human factors in computing systems*. 2013.
73. Benton, A., M. Mitchell, and D. Hovy, *Multi-task learning for mental health using social media text.* arXiv preprint arXiv:1712.03538, 2017.
74. Badian, Y., et al., *A Picture May Be Worth a Thousand Lives: An Interpretable Artificial Intelligence Strategy for Predictions of Suicide Risk from Social Media Images.* arXiv preprint arXiv:2302.09488, 2023.
75. Shen, R., et al. *Depression detection by analysing eye movements on emotional images*. in *ICASSP 2021-2021 IEEE International Conference on Acoustics, Speech and Signal Processing (ICASSP)*. 2021. IEEE.
76. Zhou, X., et al., *Visually interpretable representation learning for depression recognition from facial images.* IEEE transactions on affective computing, 2018. **11**(3): p. 542-552.
77. Rudin, C., *Stop explaining black box machine learning models for high stakes decisions and use interpretable models instead.* Nature machine intelligence, 2019. **1**(5): p. 206-215.
78. Hatherley, J., R. Sparrow, and M. Howard, *The virtues of interpretable medical artificial intelligence.* Cambridge Quarterly of Healthcare Ethics, 2022: p. 1-10.
79. Ahmad, M.A., C. Eckert, and A. Teredesai. *Interpretable machine learning in healthcare*. in *Proceedings of the 2018 ACM international conference on bioinformatics, computational biology, and health informatics*. 2018.
80. Custode, L.L. and G. Iacca. *Interpretable AI for policy-making in pandemics*. in *Proceedings of the Genetic and Evolutionary Computation Conference Companion*. 2022.

81. Antoniou, G., E. Papadakis, and G. Baryannis, *Mental health diagnosis: a case for explainable artificial intelligence.* International Journal on Artificial Intelligence Tools, 2022. **31**(03): p. 2241003.
82. Joyce, D.W., et al., *Explainable artificial intelligence for mental health through transparency and interpretability for understandability.* npj Digital Medicine, 2023. **6**(1): p. 6.
83. Ming, Y., H. Qu, and E. Bertini, *Rulematrix: Visualizing and understanding classifiers with rules.* IEEE transactions on visualization and computer graphics, 2018. **25**(1): p. 342-352.